\documentclass[lettersize,journal]{IEEEtran}
\usepackage{amsmath,amsfonts}
\usepackage{algorithm}
\usepackage{array}
\usepackage[caption=false,font=normalsize,labelfont=sf,textfont=sf]{subfig}
\usepackage{textcomp}
\usepackage{stfloats}
\usepackage{url}
\usepackage{verbatim}
\usepackage{graphicx}
\usepackage{cite}
\usepackage{multirow}
\usepackage{booktabs}

\usepackage{algpseudocode}

\begin{document}

\title{EAGLE: Towards Efficient Arbitrary Referring Visual Prompts Comprehension for Multimodal Large Language Models}

\author{Jiacheng~Zhang,
        Yang~Jiao,
        Shaoxiang~Chen,
        Jingjing~Chen,
        and~Yu-Gang~Jiang
\thanks{Jiacheng Zhang, Yang Jiao, Jingjing Chen, and Yu-Gang Jiang are with Fudan Vision and Learning Laboratory (FVL), Fudan University, Shanghai, China. Shaoxiang Chen is with Meituan.}}

\markboth{IEEE TRANSACTIONS ON MULTIMEDIA}%
{Shell \MakeLowercase{\textit{et al.}}: A Sample Article Using IEEEtran.cls for IEEE Journals}



\maketitle

\begin{abstract}
Recently, Multimodal Large Language Models (MLLMs) have sparked great research interests owing to their exceptional content-reasoning and instruction-following capabilities. To effectively instruct an MLLM, in addition to conventional language expressions, the practice of referring to objects by painting with brushes on images has emerged as a prevalent tool (referred to as \textbf{\emph{"referring visual prompts"}} in this paper) due to its efficacy in aligning the user's intention with specific image regions. 
To accommodate the most common referring visual prompts, namely points, boxes, and masks, existing approaches initially utilize specialized feature encoding modules to capture the semantics of the highlighted areas indicated by these prompts. Subsequently, these encoded region features are adapted to MLLMs through fine-tuning on a meticulously curated multimodal instruction dataset. However, such designs suffer from redundancy in architecture as they overlook the inherent region-level comprehension capabilities of MLLMs. 
Moreover, they face challenges in effectively generalizing when encountering a diverse range of arbitrary referring visual prompts in real-life scenarios, primarily due to their sensitivity to the quality of the provided referring visual prompts. 
To address the above issues, we propose \textbf{EAGLE}, a novel MLLM that empowers comprehension of arbitrary referring visual prompts with less training efforts than existing approaches. 
Specifically, our EAGLE maintains the innate format of the referring visual prompts as colored patches rendered on the given image for conducting the instruction tuning. 
Unlike previous methods that focus on initiating region-text alignment in semantics, our approach embeds referring visual prompts as spatial concepts conveying specific spatial areas comprehensible to the MLLM, with the semantic comprehension of these regions originating from the MLLM itself. Besides, we also propose a Geometry-Agnostic Learning paradigm (GAL) to further disentangle the MLLM's region-level comprehension with the specific formats of referring visual prompts. Extensive experiments are conducted to prove the effectiveness of our proposed method.
\end{abstract}


\begin{figure*}
    \centering
    \includegraphics[width=0.9\textwidth]{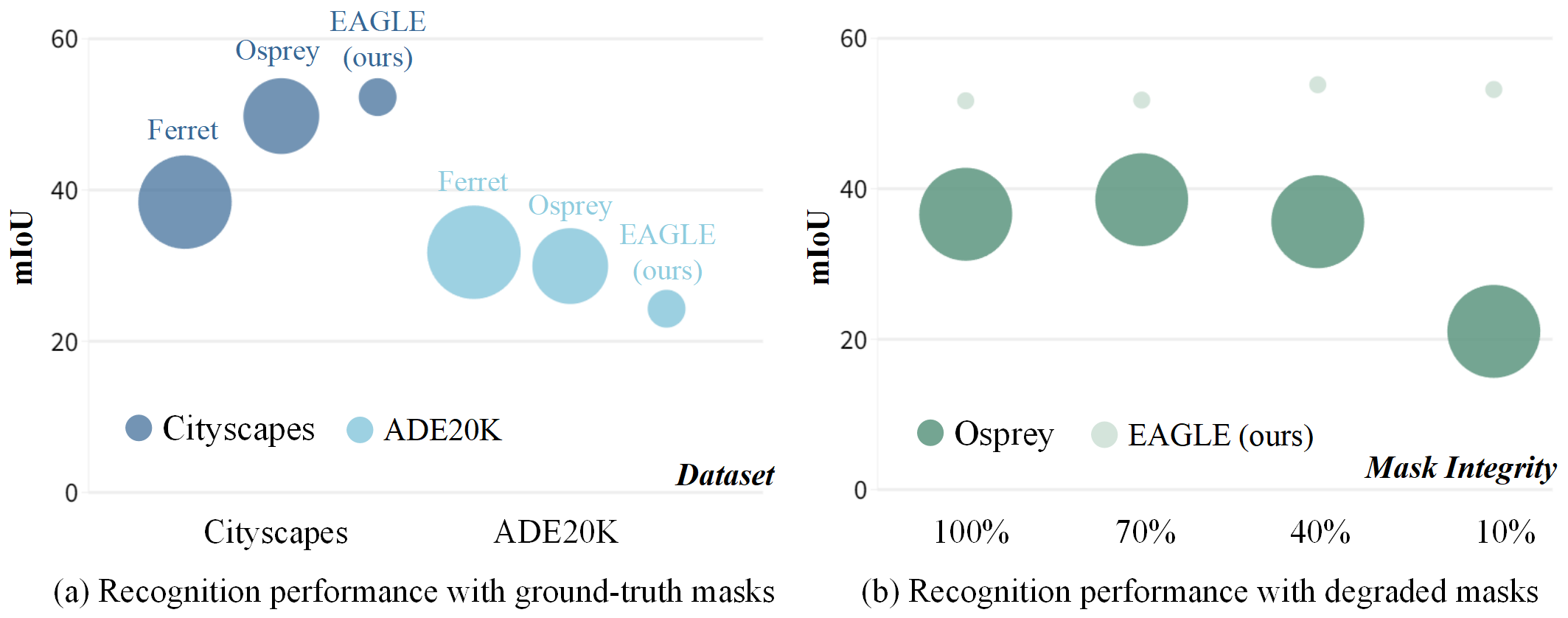}
    \caption{Performances comparison of prevalent MLLMs and our proposed EAGLE on (a) different datasets and (b) arbitrary referring visual prompts. We leverage semantic segmentation as the proxy task to evaluate the referring visual prompt recognition capabilities of MLLMs following Osprey~\cite{yuan2023osprey}. In (a), we follow the configuration of Osprey by simply using ground-truth masks as the visual prompt. 
    In (b), we imitate arbitrary referring visual prompts by degenerating the integrity of ground-truth masks, in which we demonstrate the consistent performance of our GAL paradigm in dealing with arbitrary referring visual prompts. The area of circles indicates the scale of employed training data.} 
    \label{fig:fig1}
\end{figure*}

\section{Introduction}
\IEEEPARstart{M}{ultimodal}
 Large Language Models (MLLMs)~\cite{bai2023qwen,peng2023kosmos,zhu2023minigpt,jiao2024lumen,dai2024instructblip,alayrac2022flamingo} have recently emerged as a hot research topic within the realm of computer vision. Inheriting the remarkable reasoning and instruction-following capabilities of Large Language Models (LLMs), MLLMs take a step forward by connecting the visual world with the language space, thereby vastly expanding their potential applications across various disciplinary fields.~\cite{yin2023foodlmm,wang2023huatuo,jiao2024lumen}. 

To align the vision and language modalities, existing MLLMs~\cite{bai2023qwen,zhu2023minigpt,liu2024visual} prominently leverage image-caption pairs by prompting the MLLM with manufactured instructions and a given image, then training the model to generate the captions corresponding to this image. To further augment the region-level comprehension capability, several studies~\cite{peng2023kosmos,chen2023shikra,zhang2023gpt4roi,you2023ferret,yuan2023osprey} propose to extend the MLLM's instruction repertoire with \textbf{\emph{``referring visual prompts''}}, offering users an interface to refer the detailed regions within the provided image. Kosmos-2~\cite{peng2023kosmos} and Shikra~\cite{chen2023shikra} facilitate the use of bounding boxes as referring visual prompts by incorporating their coordinates into user instructions. To accommodate a broader range of referring visual prompt formats such as points and regional masks, Ferret~\cite{you2023ferret} and Osprey~\cite{yuan2023osprey} employ diverse encoding techniques to extract region features. These features are subsequently mapped into the MLLM's input space through fine-tuning on the meticulously curated instruction datasets. However, despite their impressive regional comprehension capabilities, both Ferret and Osprey suffer from two major drawbacks:

(1) \textbf{Redundant in architecture}. The cooperation of regional feature encoding modules and curated instruction datasets is leveraged to endow the MLLMs with region-level comprehension capability, however, this is redundant because the MLLM naturally possesses such a capability during the vision-language alignment learning using paired image-caption datasets. This stems from the MLLM's inherent requirement to perceive objects and comprehend their relationships within a given image to generate accurate captions. Therefore, we posit that the crux of unleashing the region-level comprehension potential of MLLMs is to efficiently embed the referring visual prompts as the concepts conveying certain spatial areas comprehensible to MLLMs, rather than initiating region-text alignment from scratch with the aid of a curated large-scale dataset.

(2) \textbf{Poor generalizable in function}. They maintain an ideal running configuration, utilizing referring visual prompts with consistent formats both during training and inference. For example, Osprey \cite{yuan2023osprey} employs the ground-truth boxes and masks as referring visual prompts. However, this is intractable in practice, as users lacking professional skills may draw arbitrary shapes and formats of referring visual prompts, often differing significantly from those used during training. Therefore, when generalizing to real-world scenarios, Osprey will degrade drastically as shown in Fig~\ref{fig:fig1}(b).

To address the aforementioned problems, in this paper, we propose EAGLE, a novel MLLM supporting efficient comprehension of arbitrary referring visual prompts. Our EAGLE comprises two pivotal designs. First of all, we render diverse formats of referring visual prompts in the form of colored patches onto the image, which will serve as image resources of instructional datasets for performing the instruction tuning. This design not only minimally modifies existing MLLMs but also respects their innate region-level comprehension capabilities. As demonstrated in Fig~\ref{fig:fig1}(a), even with less training data, our design can achieve comparable performances to both Ferret and Osprey, which proves that properly prompting the MLLM to recognize detailed areas is more efficient than initiating a new region-text alignment process. Secondly, we propose a Geometry-Agnostic Learning paradigm (GAL) to deal with arbitrary shapes and formats of referring visual prompts. We analyze that the degraded performance of Ferret and Osprey when handling arbitrary referring visual prompts stems from the incomplete or noisy object semantics due to the unsatisfactory coverage of these prompts. Toward this end, our GAL disentangles the region-level recognition with referring visual prompt geometry by reformulating diverse referring visual prompts into a set of representative points uniform in formats. By further employing these reformulated representative points for the aforementioned instruction tuning, the MLLM is encouraged to focus on the primarily referred object, regardless of the shapes or formats of the referring visual prompts. 
Benefiting the collaboration of these two novel designs, our EAGLE model can efficiently handle arbitrary referring visual prompts more effectively than previous state-of-the-art approaches as shown in Fig~\ref{fig:fig1}(b).

In general, our contributions lie in three-fold:
\begin{itemize}
\item We propose EAGLE, a novel MLLM that empowers the comprehension capacity of arbitrary referring visual prompts by enhancing the local information of original image features. Our method avoids the introduction of additional region-encoding modules and requires less training effort.
\item We propose Geometry-Agnostic Learning (GAL), a paradigm that supports the disentanglement of referring visual prompts with diverse shapes and formats. GAL achieves the degradation and unification of region annotations, which alleviate the influence of shapes and formats.
\item To assess the performance of our approach on diverse referring visual prompts, we propose a generation method to craft human-drawn style masks for further evaluation. Experiments show that our proposed EAGLE can effectively deal with irregular region annotations.
\end{itemize}

\begin{figure*}[h]
    \centering
    \includegraphics[width=\textwidth]{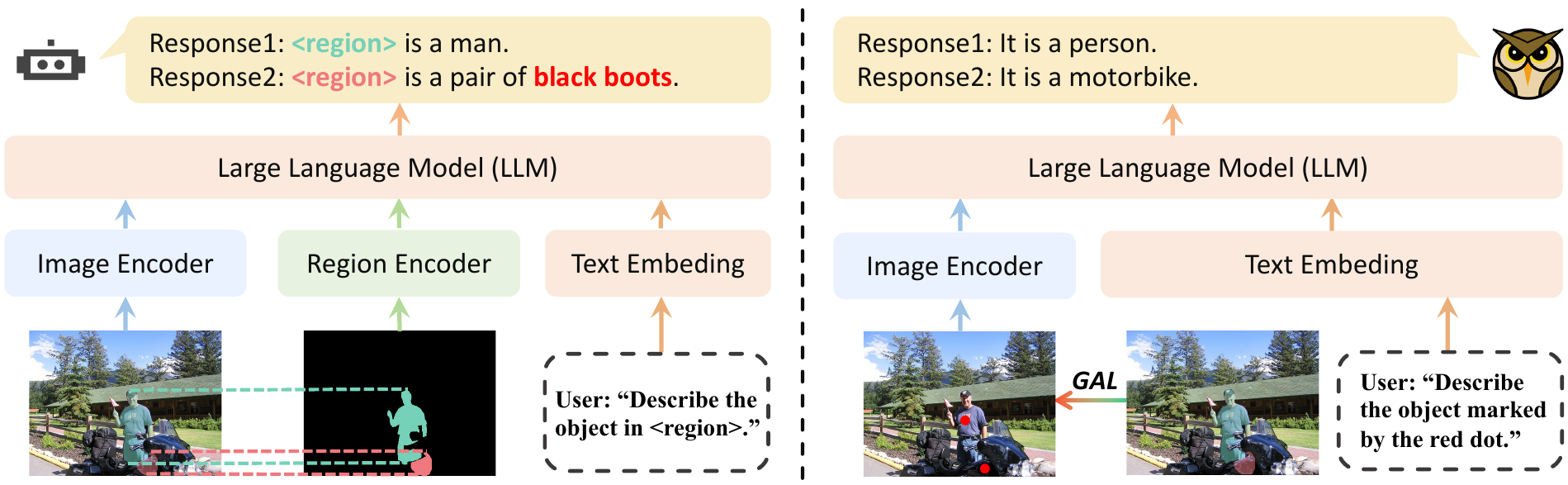}
    \caption{Comparison of our proposed Eagle with previous methods. Previous approaches (left column) specifically extract region features and project them into LLM's input space. Our EAGLE (right column) preprocesses arbitrary referring visual prompts with the introduced  Geometry-Agnostic Learning (GAL) paradigm, and then renders them onto the image for highlighting the referred regions.}
    \label{fig:fig2}
\end{figure*}

\section{Related Works}
In this section, we give a brief introduction to multimodal large language models, region-level image understanding, and visual prompt engineering.

\subsection{Multimodal Large Language Models.} 
In natural language processing (NLP), Large Language Models (LLM) like GPT-3 \cite{floridi2020gpt3}, PaLM \cite{chowdhery2023palm}, LLaMA \cite{touvron2023llama}, Vicuana \cite{zheng2024vicuna}, and Guanaco \cite{dettmers2024qlora} demonstrated strong capacities across a wide range of language tasks. The rapid advancement of LLMs has also propelled the progress of MLLMs. Recent MLLMs, like LLaVA \cite{liu2024visual}, MiniGPT-4 \cite{zhu2023minigpt}, Qwen-VL \cite{bai2023qwen}, and Video-LLaMA \cite{zhang2023video}, have been proposed. They shared a similar architecture, with a vision encoder to extract image features, an adaptor to project image features onto textual embedding spaces, and an LLM to process both text and projected image features.
This unified and concise model structure has spawned a new instruction-based task framework. To enable MLLM to accomplish the tasks described by the user's instructions, visual instruction tuning is adopted in these works.
The central part of visual instruction tuning lies in the construction of instruction-following data. Specifically, they converted existing datasets into dialog format with the help of LLMs like GPT-4 \cite{achiam2023gpt4}, which typically contain diverse QA about image properties, detailed image descriptions, and complex logical reasoning based on the image information. After training on these curated datasets, MLLMs showed remarkable performance on user-provided arbitrary task instructions.

\subsection{Region-level Image Understanding.}
Instruction tuning on image-text pairs curated data endows MLLMs with global comprehension capabilities, yet it proves inadequate for tasks requiring detailed region understanding. Thus, works ~\cite{peng2023kosmos,chen2023shikra,you2023ferret,yuan2023osprey,zhang2023gpt4roi}  have been proposed to facilitate region-level understanding. For the region format, early works utilized points and bounding boxes as the region indicator. Kosmos-2 \cite{peng2023kosmos} used location tokens to represent bounding boxes for referring and grounding tasks. Shikra \cite{chen2023shikra} directly represents both points and bounding boxes in natural language form, without using extra modules. We share similarity with Shikra. But instead of describing points in natural language, we render points on the image. In the visual space, points exhibit more flexibility with color, radius, transparency, and even shape.

Since the bounding box contains redundant and inaccurate information, recent works have leveraged masks as the region indicator for meticulous region-text alignment. Ferret \cite{you2023ferret} initially attempted to cope with free-shape regions, especially masks. They proposed a spatial-aware visual sampler to extract features on variable referred regions in a PointNet++ manner. Furthermore, Osprey \cite{yuan2023osprey} constructed a 724k mask-text instruction dataset and tuned their model for fine-grained pixel-level image understanding. However, they rely on the shape of masks to extract regional features, which affects their performance on irregular regions. On the contrary, we propose a region degradation method to circumvent the impact of variable shapes. 
% Compared with bounding boxes, our point-rendering method serves as a summarization of regional information.
Besides, we set up a new benchmark to bridge the gap in evaluating the performance of MLLMs against incomplete, variably-shaped masks.

\subsection{Visual Prompt Engineering}
Prompt tuning has gained widespread popularity and adoption across both natural language processing ~\cite{liu2021ptuning, lester2021power, gao2020making} and computer vision communities ~\cite{tsimpoukelli2021frozen, yao2024cpt, jia2022visualprompttuning, bahng2022exploring}. Common practices include learning task-specific tokens \cite{jia2022visualprompttuning} or patterns \cite{bahng2022exploring}. 
Different from visual prompt tuning, visual prompt engineering directly renders artificial and salient visual markers onto the image. \cite{shtedritski2023what} pioneered rendering red circles around objects to guide the model's attention. This simple operation yields substantial performance on zero-shot referring expressions comprehension and keypoint localization tasks. CPT \cite{yao2024cpt} proposed a new fine-tuning paradigm for visual grounding tasks by marking image regions with colored blocks or masks. FGVP \cite{yang2024fine} introduced the blur reverse mask as the visual prompt to facilitate zero-shot referring expression comprehension.

Furthermore, visual prompts have also been adopted in MLLMs. M3IT \cite{li2023m3it} curated a 2.4M multi-modal, multilingual instruction tuning dataset. They involved adding visible bounding boxes for datasets with region annotations to replace common text descriptions. \cite{cai2023making} proposed ViP-LLaVA to directly process images with arbitrary visual prompts.
Among all these prior works, ViP-LLaVA is most similar to us as they also enable MLLMs to understand visual prompts. However, our research exhibits several notable distinctions. First, we focus on alleviating the impact of user-drawn, irregularly-shaped regions via region degradation. We leverage visual point prompts as a specific solution since they can direct the model's attention without introducing additional modules. Second, we have established a novel benchmark by reconstructing the ground-truth masks from existing datasets into scribble format to simulate user-drawn regions. Through evaluation on our proposed benchmark, we assess the regional understanding ability of different MLLMs against incomplete, irregular masks.

\begin{figure*}[ht]
    \centering
    \includegraphics[width=0.8\textwidth]{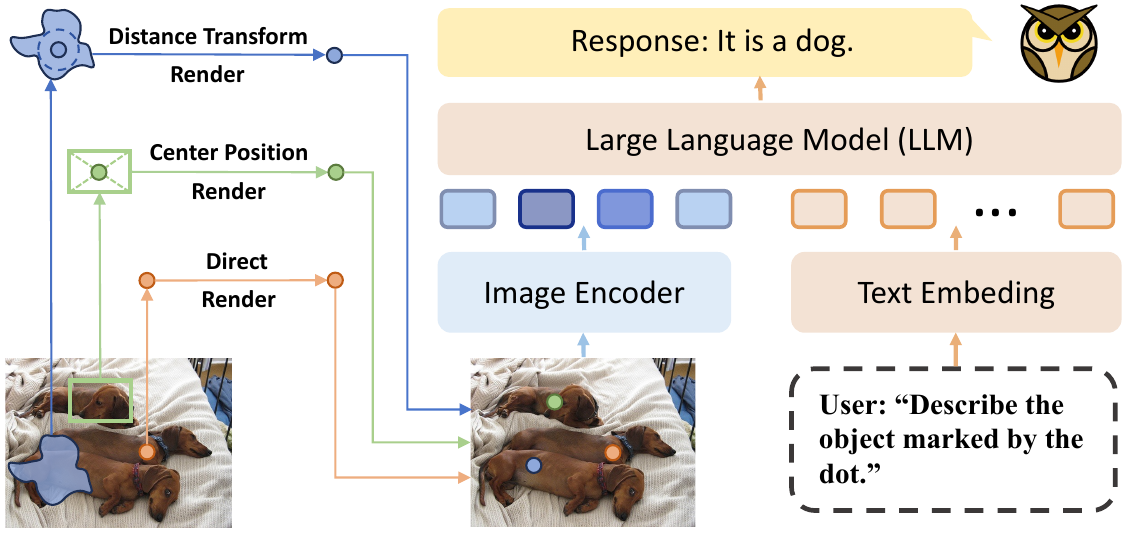}
    \caption{The overall framework of our proposed EAGLE. Given arbitrary referring visual prompts, our method first transforms them into a set of uniform points to disentangle the later region-level recognition learning with diverse shapes and formats of referring visual prompts. These transformed points are rendered onto the input image to highlight referred regions while not harming the complete semantics of the original image. Afterward, the rendered image and user instructions are fed into the MLLM to generate the final response.}
    \label{fig:framework}
\end{figure*}

\begin{figure*}
    \centering
    \includegraphics[width=\textwidth]{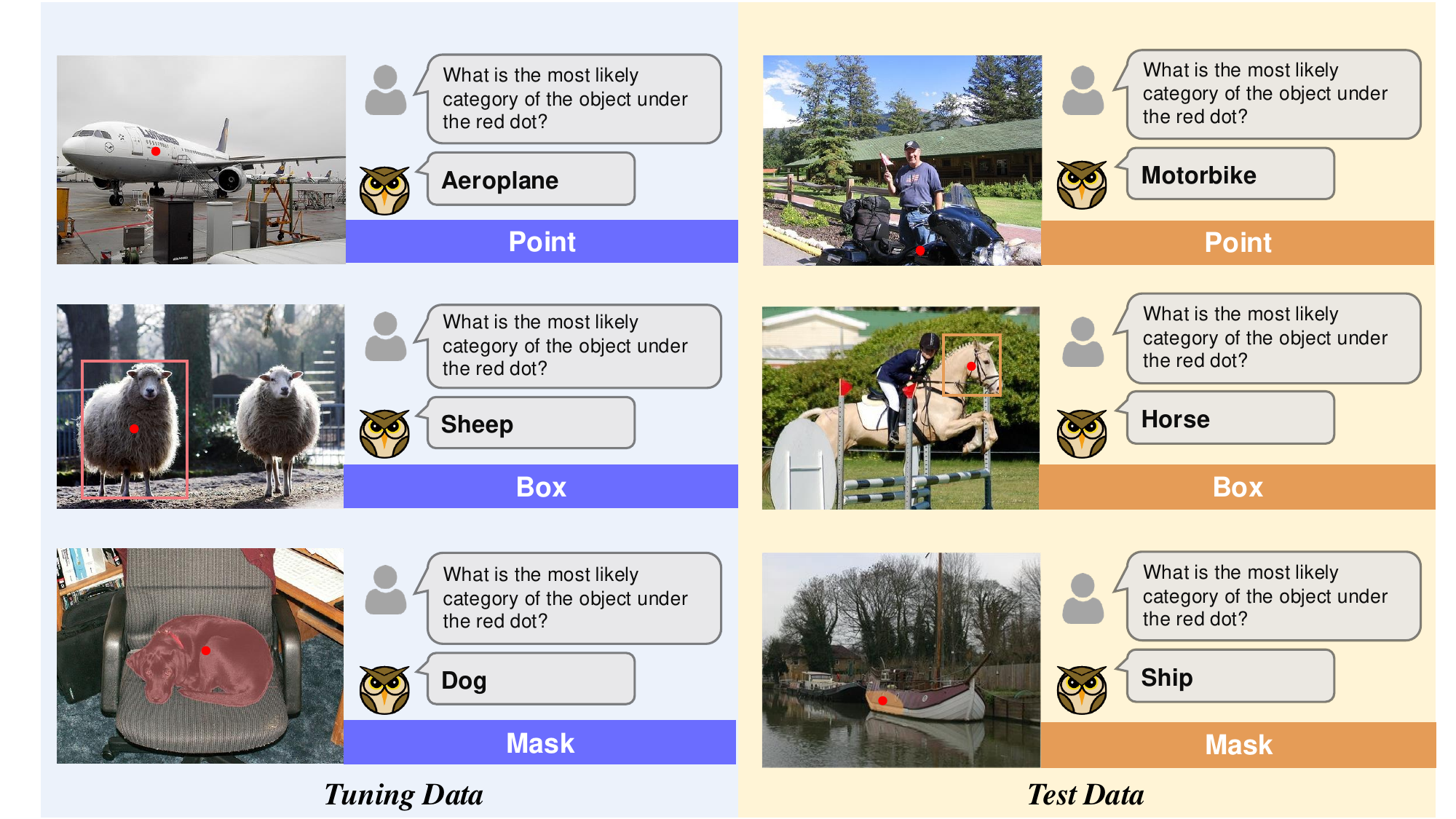}
    \caption{Illustration of reformulated conversational data with multimodal instructions. For training data, we adopt the ground-truth annotations as referring visual prompts. For test data, we degrade the original annotations to imitate the arbitrary referring visual prompts encountered in real-life scenarios.}
    \label{fig:train_and_test}
\end{figure*}

\section{Multimodal Instruction}
To facilitate MLLMs to be capable of understanding diverse formats of referring visual prompts, it is necessary to reconstruct existing multimodal datasets into the instruction following data. We focus on converting datasets with region-level annotations, including points, bounding boxes, and masks, into the human-model conversation formats to establish the accurate mapping between regions and textual descriptions. 
Specifically, we define the user instructions in a standardized way with \textit{"\textless img\textgreater  \{user instructions\} \{region prompt\}"}, where \textless img\textgreater 
denotes special tokens with image embedding features extracted from visual encoders. \{\textit{user instructions}\} represents a broad range of queries or instructions provided by users, like "Describe this object". 
Notably, different from the conventional leverage of special tokens to signify local regions, we employ distinct visible visual prompts as visual instructions to represent specific regions, and establish the correspondence between regions and visual prompts with textual descriptions \{\textit{region prompt}\}. Here, we adopt detailed, qualifying descriptions regarding color and type to precisely locate visual prompts. For instance, \{\textit{region prompt}\} can be substituted with descriptors such as "red dot," "purple circle", etc. 

For the construction of images with visual prompts, we conduct the rendering based on the region annotations.
Given the image $x_v$ and the region annotation $R$, the image with visual prompts $\hat{x_v}$ can be obtained as follows:
\begin{equation}
    \hat{x_v} = \text{VP} (x_v, R, \Theta)
\end{equation}
where $VP$ represents the operation that renders visual prompts onto the image, and $\Theta$ denotes different visual prompt formats, including dots, boxes, masks, contours, etc.

In previous methods, region annotations are typically utilized to extract local features by the additional feature encoder. These region features are further projected into the embedding space of LLM as special tokens to provide regional information. This paradigm achieves LLM's understanding of regions by aligning regional and textual features. However, visual prompts, as visual instructions represented on the image, adopt artificial annotations significantly different from the image style to guide the model's attention. Without extracting additional regional features, visual prompts highlight local information by enhancing the original image features, thus significantly reducing the cost of equipping MLLMs with regional understanding capability.

\section{Geometry-Agnostic Learning}
In this section, we provide a detailed description of our proposed Geometry-Agnostic Learning Paradigm (GAL). We first introduce our geometry disentanglement and unification methods against various shapes and formats of region annotations. Then we demonstrate our training strategies and inference workflow.

\subsection{Geometry Disentanglement}
For the impact of arbitrary shapes and formats of referring visual prompts, we focus on the three most common region annotations, i.e. points, bounding boxes, and masks. To alleviate the influence of shapes, we propose using the point, a shapeless representation, to unify diverse area-based region annotations. In the following sections, we design corresponding degradation strategies based on the region annotation forms.

\noindent\textbf{Points}
For data that already utilize points as region annotations, we do not require any additional operations to perform region degradations. We simply render distinct and colorful point prompts to the corresponding positions on images. Given the point position $p$ provided by annotation, the image with rendered point can be directly obtained as follows:
\begin{equation}
    \hat{X_v} = \text{VP} (x_v, p, \Theta_{pnt})
\end{equation}
where $\Theta_{pnt}$ denotes the selection of visual point prompts, including point-based dots, squares, crosses, etc.

\noindent\textbf{Bounding Boxes}
Bounding boxes serve as a rough description of regions, which are frequently utilized in datasets for object detection and visual grounding. A direct way to degrade bounding boxes is to render a single point at the center position $p_c$ by: 
\begin{equation}
    \hat{X_v} = \text{VP} (x_v, p_c, \Theta_{pnt})
\end{equation}
Selecting the central point as the substitution of the bounding box is a simple yet efficient method, as the main object indicated by most bounding boxes tends to overlap with the central position. This substitution is also utilized in Ferret \cite{you2023ferret}. However, there do exist some instances where the main object deviates from the centroid. These objects typically possess hollow or non-convex characteristics, thus leading to inconsistency.

To address this rare situation, we propose a vote-based rendering strategy. Specifically, we use point grids to cover the bounding box region and obtain a set of rendered images, each with a single point prompt at different grid positions. The grid-wise rendering ensures the uniform sampling of the region. Given an image $X_v$ and the bounding box $B$, the $N$ grid positions are $P \in B^{N \times 2} = \{p_1, p_2, ..., p_N\}$, a set of images $\hat{X_v}$ with point prompts can be obtained by:
\begin{equation}
    \hat{X_v} = \{ \hat{x_v} | \hat{x_v} = VP(x_v, p_i, \Theta_pnt), p_i \in P \}
\end{equation}
Subsequently, these rendered images $\hat{X_v}$ are fed into the MLLM along with user instructions to get a set of corresponding responses. These responses will be fed into the MLLM again and allow the model to determine the final answer. 

This vote-based mechanism can effectively solve the problem of deviation between the main object and the centroid of the bounding box, but it also leads to the complexity of the inference. Thus, we still prefer to directly use the centroid to replace bounding boxes.

\noindent\textbf{Masks}
Different from the bounding box, the mask is a precise region annotation. Thus, we can randomly sample a single point to replace masks. However, the randomness may impact the performance as positions at mask edges can be sampled. Intuitively, rendering the point at the mask center can better represent the underneath region. To obtain the region center $p_c$, we use distance transform to process the given binary mask $M$, and we take the pixel farthest from the mask boundary as the region center, where we render the visible point prompts. This process can be summarized as follows:
\begin{equation}
\begin{split}
    p_c &= \text{argmax } \text{DT} (M) \\
    \hat{X_v} &= \text{VP} (X_v, p_c, \Theta_{pnt})
\end{split}
\end{equation}
where $DT(\cdot)$ represents the distance transform operation, which calculates the distance of each pixel in the foreground area from the region boundary.

\subsection{Training}
To train MLLM with our reformulated geometry-disentangled instruction tuning dataset, we follow the autoregressive design of LLaVA \cite{liu2024visual}, which computes the probability of the target answers $X_a$ by:
\begin{equation}
    p\left(\mathbf{X}_{\mathrm{a}} \mid \mathbf{X}_{\mathbf{v}}, \mathbf{X}_{\text {instruct }}\right)=\prod_{i=1}^{L} p_{\boldsymbol{\theta}}\left(x_{i} \mid \mathbf{X}_{\mathbf{v}}, \mathbf{X}_{\text {instruct }}, \mathbf{X}_{\mathrm{a},<i}\right),
\end{equation}
where $X_{instruction}$ denotes the textual instruction from the user, and $X_{a,<i}$ represents all generated tokens before the current one $x_i$. During our training, we adopt the Qwen-VL-Chat-7B \cite{bai2023qwen} as our base model, while all parameters of the large language model, the visual encoder, and the vision-language adapter are fixed. We utilize Lora to perform our reformulated geometry-disentangled instruction tuning.

\subsection{Inference.}
During the inference phase, we follow the interface setting of previous methods to allow users to provide points, bounding boxes, and masks. Then, we reformulate the region input by our proposed Geometry Disentanglement method. Specifically, we first generate the image rendered with the visual point prompt based on the provided region annotation. Especially, we use the centroid to replace boxes for simplicity. Then, we use textual descriptions, like "marker by the red point", "under the red dot", to replace the region placeholder (e.g., "\textless region\textgreater", "\textless box\textgreater") in previous methods.

Furthermore, to evaluate MLLM's region-level understanding capacity, the most straightforward approach involves assessing the model's performance in recognizing segmented regions. 
Thus, to thoroughly evaluate the region comprehension capacity of models against incomplete region annotations, we propose a reconstruction algorithm to generate human-drawn-style masks. Specifically, we first choose a random point in the ground-truth mask as the start point, and then we perform the dilation in random directions to simulate human strokes. After that, we use the Gaussian blur operation to smooth shape edges. A brief outline of the mask reconstruction algorithm is shown in Algorithm 1. Our qualitative results are also shown in Fig~\ref{fig:qual}. Our regenerated masks can effectively simulate human-drawn regions, which supports further evaluation.

\begin{algorithm}
\caption{Human-Drawn Style Mask Generation Method}
\begin{algorithmic}[1]
\Require A binary mask $M$ representing the region of interest; the size range of the dilation kernel $s$; number of iterations for dilation $T$; the start position $P$.
\Ensure Dilation-transformed mask $M'$.

\State Randomly initialize $p$
\State $\hat{M}_0 = P$
\For{$t = 0, 1, ..., T-1$}
    \State Sample a random kernel size $s' = Random(0, s) * 2+1$
    \State Sample a random dilation direction $d = Random(0, 7)$
    \State Obtain the kernel $k = CreateKernel(s', d)$
    \State $\hat{M}_{i+1} = Dilation(\hat{M}_i, k)$
\EndFor
\State Truncate exceeding part $\hat{M} = \hat{M}_T \cap M$
\State Smooth edge $M' = GaussianBlur(\hat{M})$

\end{algorithmic}
\end{algorithm}

\begin{table*}[t]
\centering
\caption{Performance comparison of our method with state-of-the-art approaches when faced with arbitrary referring visual prompts. ``Type'' indicates the formats of referring visual prompts utilized by the model. ``VP'' represents the visual point prompt.}
\begin{tabular}{@{}lccccccccc@{}}
\toprule
\multirow{2}{*}{\textbf{Method}} &
  \multirow{2}{*}{\textbf{Type}} &
  \multicolumn{2}{c}{\textit{With background category}} &
  \multicolumn{5}{c}{\textit{Without background category}} &
  \multirow{2}{*}{\textbf{Avg.}} \\ \cmidrule(lr){3-9}
           &      & VOC21 & Context60 & VOC20 & City. & Context59 & ADE   & COCO-stf. &       \\ \midrule
Kosmos-2 \cite{peng2023kosmos}  & Box  & 15.11 & 7.02      & 23.15 & 3.30  & 7.54      & 4.43  & 4.99      & 9.36  \\
Qwen-VL-Chat-7B~\cite{bai2023qwen}         & VP   & 26.92 & 10.22 & 42.61 & 2.17 & 11.52 & 5.47 & 7.21 & {15.16} \\
Shikra-7B \cite{chen2023shikra} & Box  & 24.07 & 16.97     & 35.25 & 8.66  & 17.00     & 10.53 & 12.35     & 17.83 \\
GPT4RoI \cite{zhang2023gpt4roi}   & Box  & 40.56 & 24.44     & 64.08 & 15.54 & 25.44     & 12.21 & 15.72     & 28.28 \\
Ferret-7B \cite{you2023ferret}  & Mask & 53.58 & 21.49     & 51.89 & 12.37 & 23.56     & 12.76 & 15.41     & 27.29 \\
Osprey-7B \cite{yuan2023osprey} & Mask & 44.80 & 24.20     & 68.38 & 20.33 & 27.07     & 16.47 & 17.93     & 31.31 \\
Osprey-7B* \cite{yuan2023osprey} & Mask & 40.74 & 45.31     & 60.72 & 24.50 & 42.40     & 21.06 & 24.03     & 36.97 \\ \midrule
EAGLE (Ours)* &
  VP &
  \textbf{74.01} &
  \textbf{47.18} &
  \textbf{78.87} &
  \textbf{34.60} &
  \textbf{44.44} &
  \textbf{40.16} &
  \textbf{38.93} &
  \textbf{51.17} \\ \bottomrule
\end{tabular}
\end{table*}

\section{Experiments}
In this section, we conduct extensive experiments to show the great performance of our method when encountering incomplete mask inputs. Our thorough experiments verify the effectiveness of our proposed EAGLE.

\subsection{Semantic Segmentation}
\textbf{Experiment Settings} We select and reconstruct five semantic segmentation benchmarks, including PASCAL VOC 2012 \cite{pascalvoc}, PASCAL Context \cite{pascalcontext}, COCO-Stuff \cite{cocostuff}, Cityscapes \cite{cityscapes} and ADE20k \cite{ade20k}. To thoroughly evaluate different methods, we especially considered the background category, which followed the dataset setting of \cite{wang2023sclip}. The original dataset with background category is denoted as VOC21, Context60, while VOC20, Context59 denote the version without considering the background category. COCO-Stuff, ADE20k, and Cityscapes are also divided into the group without background category.

In order to fairly evaluate the performance of different methods when facing inconsistency in mask integrity between the model training and inference phases, we conducted the experiments with our regenerated human-drawn style masks based on the validation set of the above datasets. We choose Qwen-VL-Chat \cite{bai2023qwen}, Kosmos-2 \cite{peng2023kosmos}, Shikra \cite{chen2023shikra}, GPT4RoI \cite{zhang2023gpt4roi}, Ferret \cite{you2023ferret}, Osprey \cite{yuan2023osprey} as our baseline comparison methods. We first test all these methods in the open-vocabulary setting to show their performance against incomplete masks. Furthermore, we choose the best-performing method Osprey as our comparison method in a supervised setting. 

For the evaluation metric, we calculate the mIoU. Since we have reconstructed masks of the validation sets, the ground-truth masks are reset accordingly for the precise assessment. We follow the setting in Osprey that uses Sentence-Bert \cite{reimers2019sentence} to calculate the semantic similarity between the model's response and the category list of each dataset. We adopt the category name with the highest semantic similarity score as the prediction.

\textbf{Training Details}
We adopt Qwen-VL-Chat-7B \cite{bai2023qwen} as our base model for the supervised semantic segmentation evaluation. We fine-tuneQwen-VL-Chat-7B on training sets of PASCAL VOC 2012 \cite{pascalvoc}, PASCAL Context \cite{pascalcontext}, COCO-Stuff \cite{cocostuff}, Cityscapes \cite{cityscapes} and ADE20k \cite{ade20k} respectively by utilizing instruction tuning. To facilitate Qwen-VL-Chat-7B to understand visual point prompts, we render one single distinct point at the centroid of ground-truth masks. Our instruction dataset is assembled into the basic QA format, where the question is established as "What is the category of the object under the \textless color\textgreater \textless prompt form\textgreater?", while the answer is directly set to the category name of the marked object for simplicity. In this evaluation, we use the red round dot as our visual prompt form.

For the training of Osprey \cite{yuan2023osprey}, We adopt a similar training strategy with our EAGLE by utilizing instruction tuning. We use the final-stage version as the base model of Osprey, which is end-to-end fine-tuned on their curated Osprey-724K dataset. And we use the default short-form prompt of Osprey to construct the instruction tuning data with the basic QA format, where the question is "What is the category of the object in \textless mask\textgreater? Answer the question using a single word or phrase." and the answer is also the category name like our EAGLE.

We utilize the AdamW \cite{loshchilov2016sgdr} as the optimizer and the cosine annealing scheduler \cite{loshchilov2017decoupled} to adjust the learning rate. Especially, we use Lora \cite{hu2021lora} to fine-tune our method since we do not introduce extra vocabulary. We set the batch size to 1 and fine-tune the model for five epochs as the default settings of Qwen-VL-Chat-7B \cite{bai2023qwen} with the learning rate of 1$e-$5. For the training of Osprey \cite{yuan2023osprey}, we follow their default training settings.

\textbf{Results}
Table 1 summarizes the main results of our methods and all comparison methods. To facilitate our EAGLE to recognize the object marked by incomplete masks, we use "What is the category of the object in \textless region\textgreater?" as the text instruction. In the meantime, we use the default text instruction designed by each baseline method in our experiment. We have three annotation types to represent regions, including bounding boxes, masks, and visual point prompts. We first evaluate our base model, i.e. Qwen-VL-Chat-7B \cite{bai2023qwen}. It can be seen that Qwen-VL-Chat exhibits poor performance when provided visual point prompt without fine-tuning. However, after fine-tuning with the visual point prompt rendered data and using the visual prompt as the region indicator, our method gains large improvement across all datasets, which demonstrates that visual point prompts are a practical and efficient method to equip MLLMs with region-level understanding ability.

Among all comparison methods, Osprey stands out with better robustness against incomplete masks. Thus, we fine-tune both Osprey and our method. As shown in the last two rows of Table 1, although the fine-tuned Osprey has shown some improvement on incomplete masks compared to its performance before training, it still falls behind our method across all datasets, especially on VOC, Cityscapes, and ADE20K. This suggests that our Geometry-Agnostic Learning paradigm can effectively erase the impact of region shapes and integrity.

\begin{figure}
    \centering
    \includegraphics[width=0.45\textwidth]{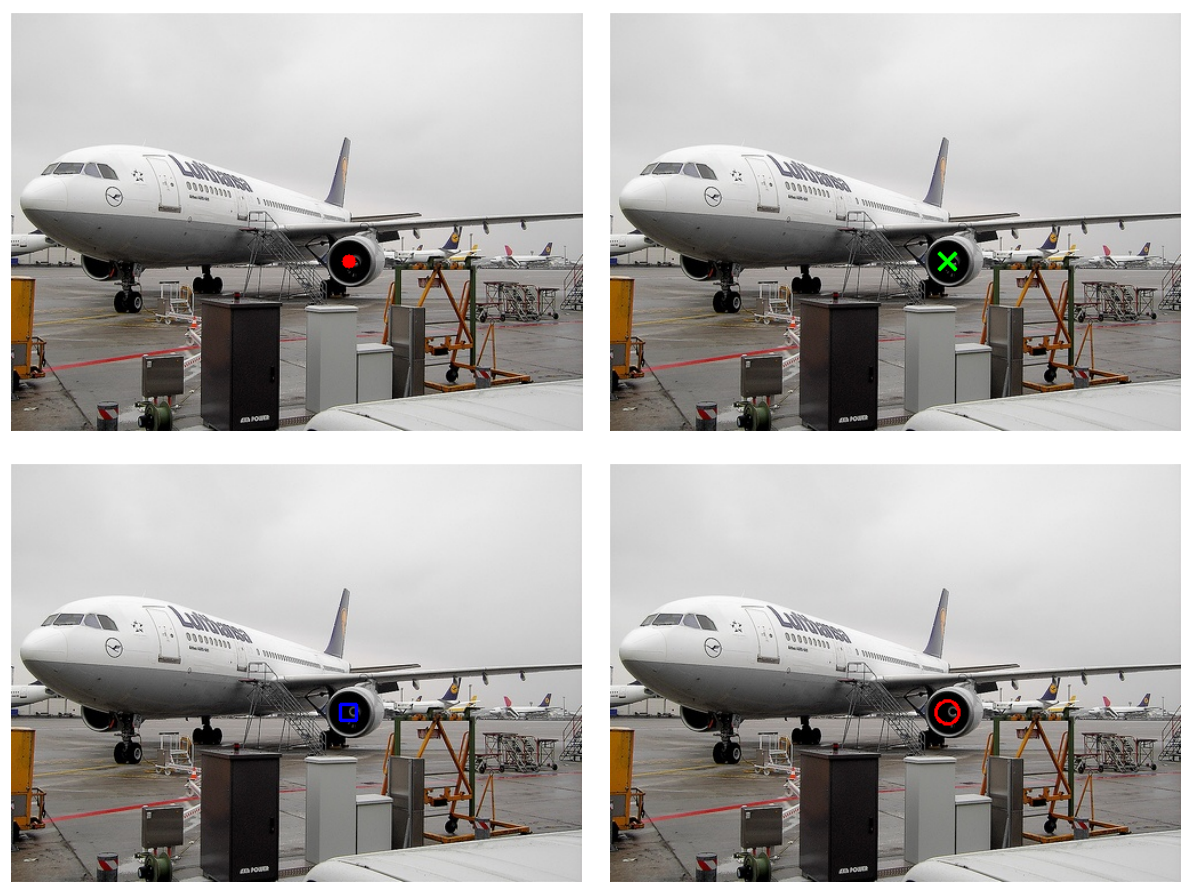}
    \caption{Illustration of different formats of rendered point markers. We evaluate the sensitivity of the MLLM to three colors (red, green, and blue) and four prompt types (dot, square, box, and circle), for a total of 12 combinations.}
    \label{fig:fig4}
\end{figure}

\begin{figure*}[ht]
    \centering
    \includegraphics[width=0.9\textwidth]{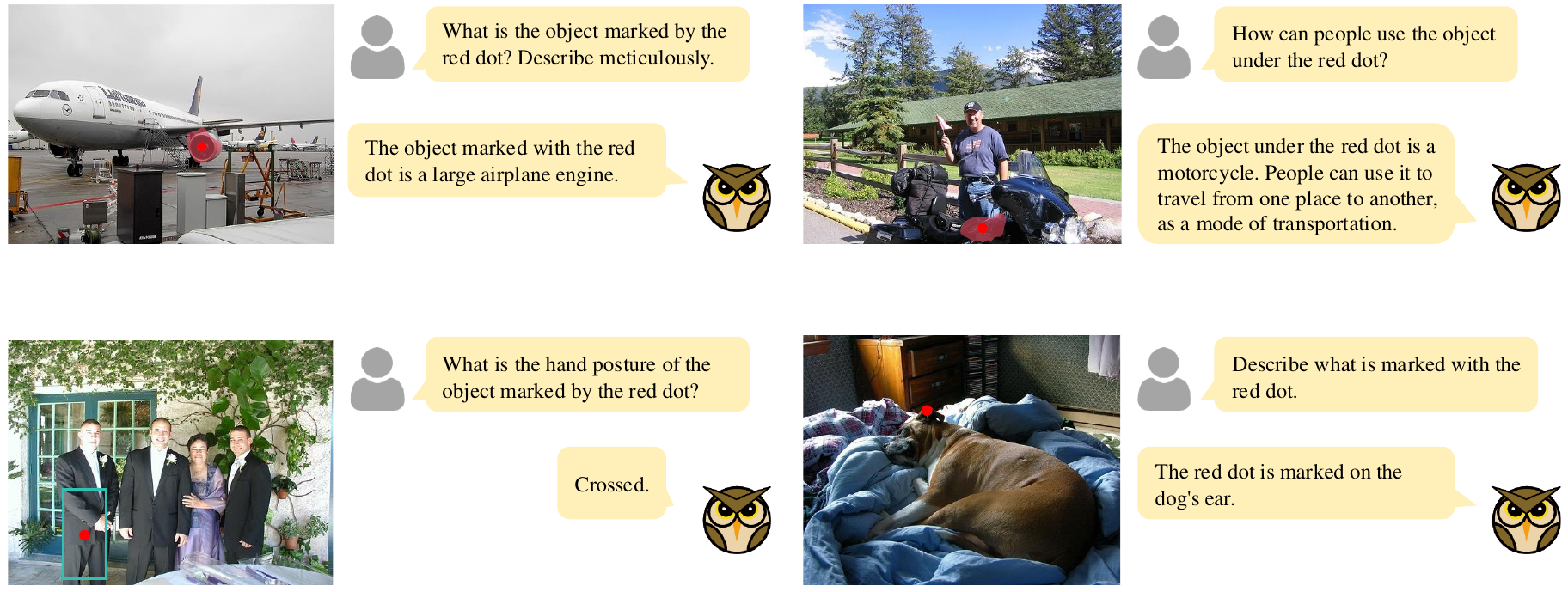}
    \caption{Qualitative results of our proposed EAGLE when faced with arbitrary referring visual prompts and user instructions beyond the scope of our training data.}
    \label{fig:qual}
\end{figure*}

\subsection{Performance on Arbitrary Boxes}
We further evaluate our method on partial boxes as arbitrary visual prompts. We conduct the experiment based on the validation set of the Pascal VOC \cite{pascalvoc} and COCO \cite{cocostuff}. We fine-tune our EAGLE following the similar setting used in our experiments on semantic segmentation tasks. For the evaluation, to simulate the partial boxes with a human-drawn style, we reconstruct the ground-truth bounding boxes in validation sets. Specifically, we randomly decrease the height and width until the area of the new box is reduced to 10\% of the ground-truth one. Furthermore, we randomly choose the position of the newly decreased box within the ground-truth bounding box to enhance the arbitrariness while keeping semantic consistency.

We keep the same setting for text instructions with slight modifications of the text instruction that may occur as the adaptation to different MLLMs. For the evaluation metric, we adopt the recognition accuracy since the reconstructed boxes are provided for region comprehension. We also use Sentence-Bert \cite{reimers2019sentence} to calculate the semantic similarity between the model's response and the category list. The category name with the highest semantic similarity is chosen as the result. We choose Ferret \cite{you2023ferret} and Osprey \cite{yuan2023osprey} as our comparison methods. Especially, we transform the box into the format of the mask for the normal inference of Osprey. As Table \ref{tab:boxes_performance} shows, our basic strategy that replaces boxes by rendering visual point prompts at centroids surpasses Ferret \cite{you2023ferret} and Osprey \cite{yuan2023osprey} on both VOC and COCO when dealing with arbitrary boxes. What is particularly noteworthy is that COCO is also utilized in Osprey during their training phase.

We further verify the effectiveness of our vote-based rendering strategy. Specifically, we render visual point prompts at the top left, bottom left, top right, bottom right, and center positions respectively. These five images are then fed into EAGLE with the same user instruction to obtain a set of responses. Finally, we adopt a standard text prompt "Here is a list of responses: [\textit{response list}]. The instruction from the user is \{\textit{user instruction}\} Please summarize these responses and answer the instruction from the user." to guide EAGLE to summarize responses and answer the user's question. As the result shows in Table \ref{tab:boxes_performance}, our vote-based mechanism outperforms the basic centroid rendering strategy.

\subsection{Ablation Study}
\textbf{Effects of the proposed GAL.}
We ablate the effects of the proposed GAL strategy in Table \ref{tab:domain}. For the configuration of ``w/o GAL'', we directly render the region mask onto the image for training and inference. From the results, it is evident that our GAL is unaffected by the discrepancy in referring visual prompts between the training and evaluation formats, demonstrating the effectiveness of our designed GAL.

\textbf{Effects of Different Color and Point Forms.}
To assess the sensitivity of MLLMs to visual point prompts with different colors and forms, we perform evaluations on VOC21, Context60, and Cityscapes datasets. We have set three colors (red, blue, and green) and four shapes (round dot, circle, square, and cross), a total of twelve combinations in our evaluation. For every combination, we use "What is the most likely category of the object under the \textless color\textgreater \textless form\textgreater?" as our text instructions, and we fine-tune the model on each dataset respectively. The results are shown in Table \ref{tab:color}. We find that the green color stands out with the highest performance across almost all datasets. We consider this may be due to the fact that the green color tends to form a more distinct contrast with other colors frequently appearing in images. 

Regarding the form of visual prompts, the model does not exhibit an obvious preference for these four types of point-based visual prompts. This indicates that MLLMs are capable of rapidly learning and understanding various forms of point-based visual prompts, thereby acquiring excellent regional understanding abilities.

\begin{table}[h]
\centering
\caption{Performance comparison with partial boxes as arbitrary referring visual prompts. We adopt recognition accuracy (\%) as our evaluation metric. "+Vote" indicates the usage of our proposed vote-based rendering mechanism.}
\label{tab:boxes_performance}
\begin{tabular}{lll}
\toprule
Method            & VOC   & COCO  \\ 
\midrule
Ferret \cite{you2023ferret}           & 36.11 & 31.65 \\
Osprey \cite{yuan2023osprey}           & 63.25 & 41.20 \\
\midrule
EAGLE* (Ours)      & 69.29 & 45.55 \\
EAGLE*+Vote (Ours) & \textbf{70.81} & \textbf{46.41} \\
\bottomrule
\end{tabular}
\end{table}

\begin{table}[h]
    % \tablestyle{10pt}{1}
    \caption{Effects of adopting different combinations of colors and prompt formats.}
    \label{tab:color}
    \centering
    \scalebox{0.9}{
    \begin{tabular}{l | c | c c c c}
        \toprule
        {Prompt Type} & {Color} & VOC21 & Cityscapes & Context60 & Avg. \\ 
        \midrule
        \multirow{3}{*}{Round dot}
             & red     & 74.01   & 34.18        & 47.18  & 51.79 \\
             & blue    & 72.28   & 29.03        & 42.30    & 47.87 \\
             & green   & 76.28   & 34.59        & 48.15  & 53.00 \\
        \midrule
        \multirow{3}{*}{Circle}
             & red     & 74.31   & 36.42        & 48.56       & 53.09 \\
             & blue    & 73.96   & 34.75        & 48.08       & 52.26 \\
             & green   & 75.71   & 39.12        & 48.63       & 54.49 \\
        \midrule
        \multirow{3}{*}{Square}
             & red     & 73.63   & 36.65        & 47.81       & 52.70 \\
             & blue    & 72.97   & 34.10        & 47.54       & 51.54 \\
             & green   & 74.81   & 37.35        & 47.80       & 53.38 \\
        \midrule
        \multirow{3}{*}{Cross}
             & red     & 74.92 & 35.74        & 48.25       & 52.97 \\
             & blue    & 75.33 & 33.55        & 48.59       & 52.49 \\
             & green   & 75.66 & 35.13        & 49.26       & 53.35 \\
        \bottomrule
    \end{tabular}}
\end{table}

\textbf{Effects of render position.}
we conduct experiments to explore the influence of rendering positions on VOC21, Cityscapes, and Context60 with our generated incomplete masks. In this evaluation, we only render one single red point onto the image. We compare our proposed method, which computes the center of the given region mask by utilizing distance transforms, with the random method that selects the rendering position randomly based on the mask. As shown in Table~\ref{tab:pos}, we find that the rendering point in the region center outperforms the random way across all datasets. 

\begin{table}[h]
\centering
\caption{Effects of the proposed GAL strategy. ``Slight'', ``Moderate'', and ``Severe'' indicate the degree of domain gap between the referring visual prompts used for training and evaluation.}
\label{tab:domain}
\begin{tabular}{lccc}
\toprule
                      & Slight  & Moderate  & Severe \\ \midrule
w/o GAL          & \textbf{73.44} & 46.08 & 11.19 \\
w GAL & 51.79 & \textbf{53.83} & \textbf{53.20} \\ \bottomrule
\end{tabular}
\end{table}

\begin{table}[htbp]
    % \tablestyle{10pt}{1}
    \caption{Effects of different point rendering strategies.}
    \label{tab:pos}
    \centering
    \begin{tabular}{l c c c}
        \toprule
        {Position}    & VOC21 & Cityscapes & Context60 \\ 
        \midrule
         Random  & 70.51   & 17.15        & 41.81      \\
         Center  & \textbf{74.01}   & \textbf{34.18}        &      \textbf{47.18}  \\
        \bottomrule
    \end{tabular}
\end{table}

\subsection{Qualitative Results}
We represent visual examples of EAGLE under multiple instructions in Fig~\ref{fig:qual}. 
EAGLE has the capacity to follow the instructions of users besides only recognizing the object category.
These visual examples demonstrate that EAGLE can understand the meaning of visual point prompts while keeping excellent conversation abilities, which verifies the effectiveness of EAGLE and our GAL paradigm.

\section{Conclusion}
In this paper, we propose EAGLE, a novel MLLM that supports the comprehension of arbitrary shapes and formats of referring visual prompts. Within our EAGLE, we introduce the Geometry-Agnostic Learning (GAL) paradigm, which conducts geometry disentanglement by reformulating arbitrary region annotation (including points, boxes, and masks) into the visual point prompts. To evaluate the performance of EAGLE, we conduct extensive experiments. The results show that EAGLE has excellent recognition performance on arbitrarily shaped regions.

\ifCLASSOPTIONcaptionsoff
  \newpage
\fi

\bibliography{IEEEfull}
\bibliographystyle{IEEEtran}

\begin{comment}
\end{comment}

\begin{IEEEbiography}[{\includegraphics[width=1in,height=1.25in,clip,keepaspectratio]{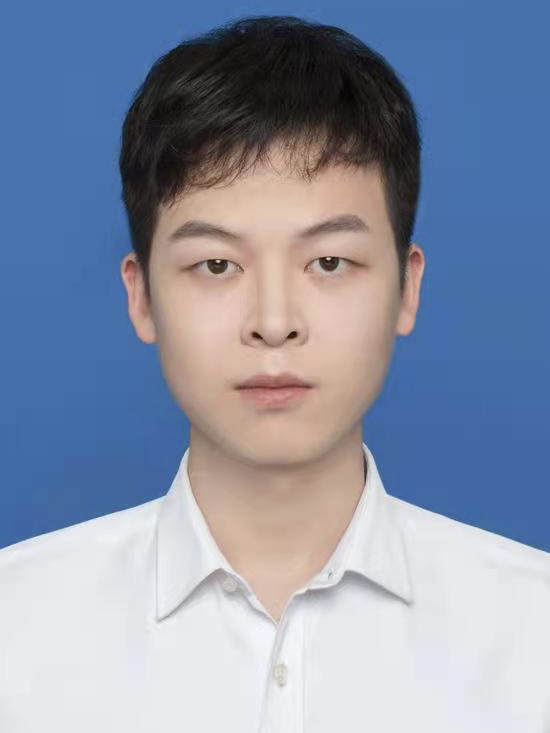}}]{Jiacheng Zhang} received the B.E. degree in Software Engineering from Wuhan University, Wuhan, China, in 2022. He is currently pursuing his M.S. degree in Computer Science at Fudan University. His research interest lies in multimodal large language models.
\end{IEEEbiography}

\begin{IEEEbiography}[{\includegraphics[width=1in,height=1.25in,clip,keepaspectratio]{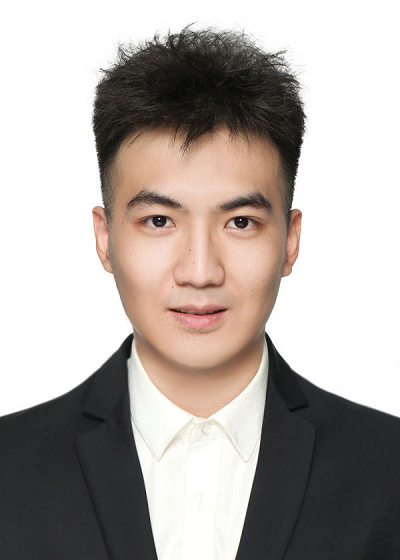}}]{Yang Jiao} received the B.E. degree from University of Electronic Science and Technology of China, Chengdu, China, in 2021. He is currently pursuing his Ph.D. degree in Computer Science at Fudan University. His research interests include multi-media analysis and 3D vision.
\end{IEEEbiography}

\begin{IEEEbiography}[{\includegraphics[width=1in,height=1.25in,clip,keepaspectratio]{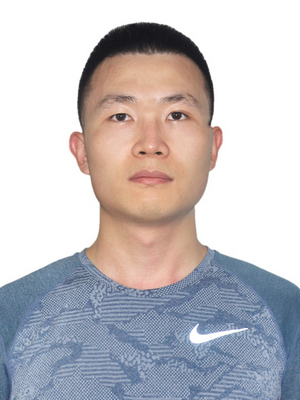}}]{Shaoxiang Chen} received the B.S. and Ph.D. degree from the School of Computer Science, Fudan University, Shanghai, China. His research interests include multimedia and deep learning, with respect to video captioning and temporal sentence localization in videos.
\end{IEEEbiography}

\begin{IEEEbiography}[{\includegraphics[width=1in,height=1.25in,clip,keepaspectratio]{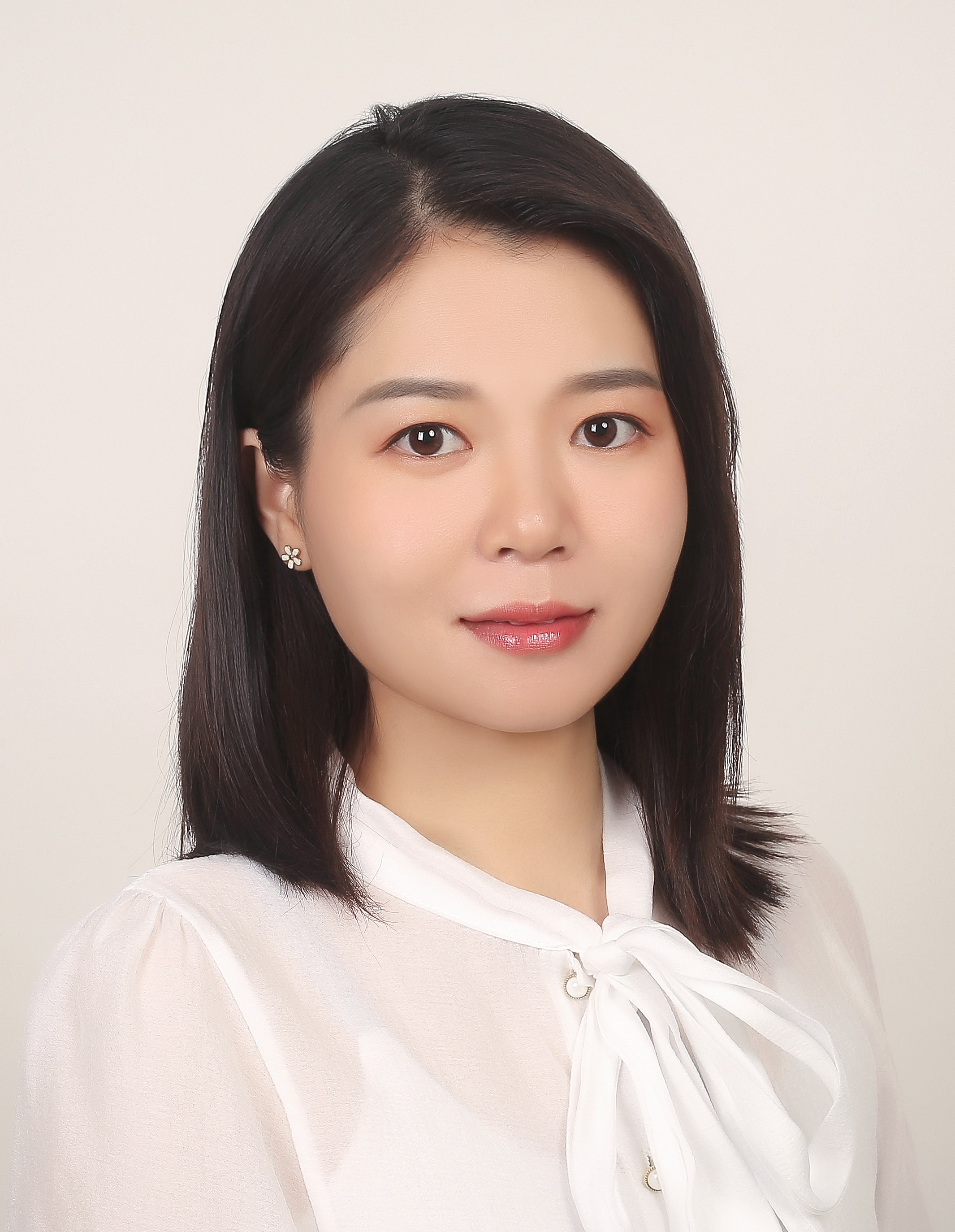}}]{Jingjing Chen} is now an Associate Professor at the School of Computer Science, Fudan University. Before joining Fudan, she was a postdoc research fellow at the School of Computing at the National University of Singapore. She received her Ph.D. degree in Computer Science from the City University of Hong Kong in 2018. Her research interest lies in diet tracking and nutrition estimation based on multi-modal processing of food images, including food recognition, and cross-modal recipe retrieval.
\end{IEEEbiography}

\begin{IEEEbiography}[{\includegraphics[width=1in,height=1.25in,clip,keepaspectratio]{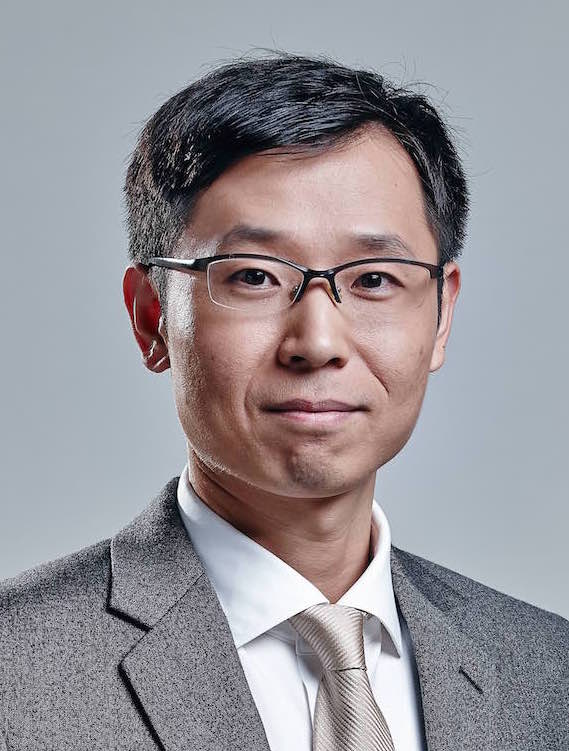}}]{Yu-Gang Jiang} received the Ph.D. degree in Computer Science from City University of Hong Kong in 2009 and worked as a Postdoctoral Research Scientist at Columbia University, New York, during 2009-2011. He is currently a Professor of Computer
Science at Fudan University, Shanghai, China. He has been elected as an IEEE Fellow in 2023. His research lies in the areas of multimedia, computer vision, and robust and trustworthy AI. His work has led to many awards, including the inaugural ACM China Rising Star Award, the 2015 ACM SIGMM Rising Star Award, the Research Award for Excellent Young Scholars from NSF China, and the Chang Jiang Distinguished Professorship appointed by the Ministry of Education of China.
\end{IEEEbiography}

\end{document}